\documentclass[11pt,a4paper]{article}
\usepackage[nohyperref]{acl2020}
\usepackage{hyperref}
\usepackage{xcolor}		
\definecolor{darkblue}{rgb}{0, 0, 0.5}
\hypersetup{colorlinks=true,citecolor=darkblue, linkcolor=darkblue, urlcolor=darkblue}
\usepackage{times}
\usepackage{latexsym}

\usepackage{microtype}

\aclfinalcopy 
\def\aclpaperid{2130} 

\usepackage{amsmath,amsfonts,bm}

\def\eqref#1{equation~\ref{#1}}

\def\1{\bm{1}}

\DeclareMathAlphabet{\mathsfit}{\encodingdefault}{\sfdefault}{m}{sl}
\SetMathAlphabet{\mathsfit}{bold}{\encodingdefault}{\sfdefault}{bx}{n}

\usepackage{booktabs}
\usepackage{cleveref}
\usepackage{graphicx}
\usepackage{enumitem}
\usepackage{float}

\usepackage[T1]{fontenc}
\usepackage[utf8]{inputenc}

\newcommand{\stories}{\textsc{Stories}}
\newcommand{\abstracts}{\textsc{Abstracts}}
\newcommand{\lyrics}{\textsc{Lyrics}}
\newcommand{\sto}{\textsc{Sto}}
\newcommand{\abs}{\textsc{Abs}}
\newcommand{\lyr}{\textsc{Lyr}}

\newcommand{\textft}[1]{{\fontfamily{lmss}\selectfont{#1}}}
\newcommand{\blank}{[\textft{blank}]}
\newcommand{\blankword}{[\textft{blank~word}]}

\newcommand{\blankdocument}{[\textft{blank~document}]}
\newcommand{\sep}{[\textft{sep}]}
\newcommand{\answer}{[\textft{answer}]}

\newcommand{\x}{x}
\newcommand{\y}{y}
\newcommand{\xtilde}{\tilde{x}}

\newcommand{\p}{p_{\theta}}
\newcommand{\Mask}{\text{Mask}(\x)}

\DeclareRobustCommand{\lm}{LM}
\DeclareRobustCommand{\lmrev}{LM-Rev}
\DeclareRobustCommand{\lmall}{LM-All}
\DeclareRobustCommand{\ilm}{ILM}
\DeclareRobustCommand{\lmscratch}{LM (scratch)}
\DeclareRobustCommand{\lmrevscratch}{LM-Rev (scratch)}
\DeclareRobustCommand{\lmallscratch}{LM-All (scratch)}
\DeclareRobustCommand{\ilmscratch}{ILM (scratch)}

\definecolor{mina-blue}{HTML}{003399}
\definecolor{mina-skyblue}{HTML}{3366CC}
\definecolor{mina-orange}{HTML}{FF6600}
\definecolor{mina-yellow}{HTML}{FF9900}

\title{Enabling Language Models to Fill in the Blanks}

\author{
Chris Donahue \\ Stanford University \\
\And Mina Lee \\ Stanford University \\ \texttt{\{cdonahue,minalee,pliang\}@cs.stanford.edu} 
\And Percy Liang \\ Stanford University \\
}

\date{}

\begin{document}

\maketitle

\begin{abstract}
We present a simple approach for \emph{text infilling}, the task of predicting missing spans of text at any position in a document. While infilling could enable rich functionality especially for writing assistance tools, more attention has been devoted to language modeling---a special case of infilling where text is predicted at the end of a document. In this paper, we aim to extend the capabilities of language models (LMs) to the more general task of infilling. To this end, we train (or fine-tune) off-the-shelf LMs on sequences containing the concatenation of artificially-masked text and the text which was masked. We show that this approach, which we call \emph{infilling by language modeling}, can enable LMs to infill entire sentences effectively on three different domains: short stories, scientific abstracts, and lyrics. Furthermore, we show that humans have difficulty identifying sentences infilled by our approach as machine-generated in the domain of short stories.
\end{abstract}

\section{Introduction}\label{sec:intro}
\begin{figure}[t!]
    \centering
    \includegraphics[width=1\linewidth]{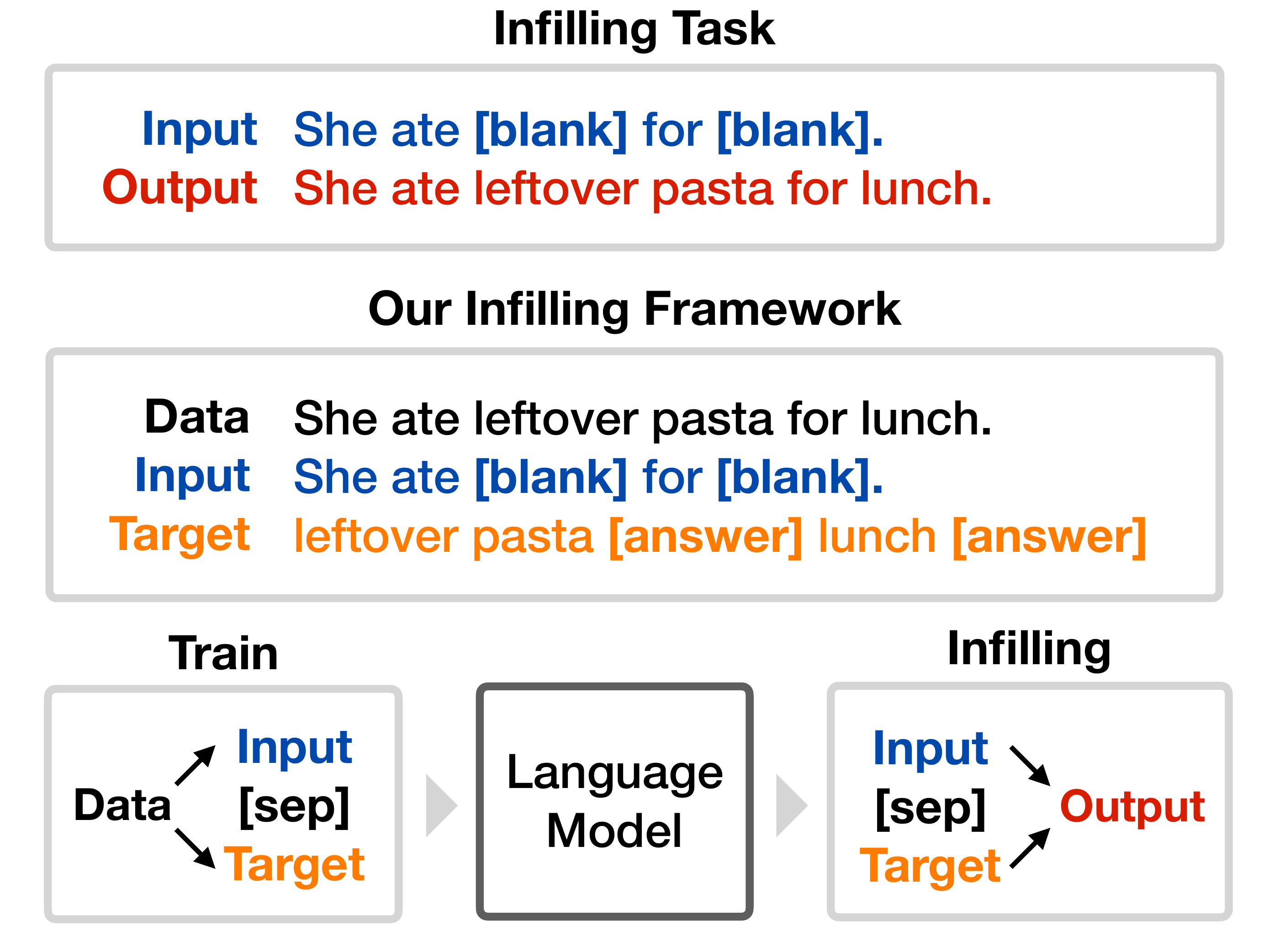}
    \caption{
    We consider the task of infilling, which takes incomplete text as input and outputs completed text.
    To tackle this task, our framework constructs training examples by masking random spans to generate pairs of inputs (text with blanks) and targets (answers for each blank). 
    We then train unidirectional language models on the concatenation of each pair.
    Once trained, a model takes text input with blanks, predicts the answers, and then combines them to produce the output.
    }
    \label{fig:infilling_training_example}
    \vspace{-5mm}
\end{figure}

Text infilling is the task of predicting missing spans of text which are consistent with the preceding and subsequent text.\footnote{Text infilling is a generalization of the \emph{cloze} task~\citep{taylor1953cloze}---cloze historically refers to infilling individual words.} 
Systems capable of infilling have the potential to enable rich applications such as assisting humans in
editing or revising text~\citep{shih2019xl}, 
connecting fragmented ideas~\citep{ai2019haim}, 
and restoring ancient documents~\citep{assael2019restoring}. 
Rather than targeting a particular application, our goal here is to provide a general, flexible, and simple infilling framework which can convincingly infill in a variety of domains.

A special case of infilling is language modeling: 
predicting text given preceding but not subsequent text.\footnote{In this paper, language modeling always refers to ordinary LMs, i.e., ``unidirectional,'' ``autoregressive,'' or ``left-to-right.''}
Language models are 
(1)~capable of generating remarkably coherent text~\citep{zellers2019defending,see2019massively},
(2)~efficient at generating text, 
and 
(3)~conceptually simple, 
but cannot infill effectively as they can only leverage context in a single direction (usually the past).
On the other hand, 
strategies such as BERT~\citep{devlin2019bert} and SpanBERT~\citep{joshi2019spanbert} are able to infill using both preceding and subsequent text. 
However, their use of bidirectional attention limits their infilling capabilities to fixed-length spans. 
This is problematic as---for many applications---we may not know the length of a missing span \emph{a~priori}. 
\citet{zhu2019text} propose a method capable of infilling variable-length spans, but it uses a specialized architecture and hence cannot easily leverage large-scale pre-trained models.

In this work, 
we present infilling by language modeling (ILM), 
a simple framework which enables LMs to infill variable-length spans while preserving their aforementioned benefits:
generation quality, efficient sampling, and conceptual simplicity. 
Our framework involves a straightforward formulation of the infilling task which, as we demonstrate, can be learned effectively by existing LM architectures. 
As shown in \Cref{fig:infilling_training_example},
our approach concatenates artificially-masked text with the text which was masked, and adopts a standard LM training (or fine-tuning) procedure on such examples. 
Once trained, infilling can be performed for a document with blanks by using the LM to generate text and then replacing the blanks with this text.

In addition to its conceptual simplicity, 
our experiments show that ILM enables off-the-shelf LMs to infill effectively. 
Furthermore, we find that infilling performance improves when starting from a large-scale pre-trained LM (as opposed to training from scratch), 
suggesting an additional benefit of using our model-agnostic framework compared to approaches which require specialized architectures.

We provide an interactive web demo of models trained under our framework. 
This demo can infill multiple variable-length spans with different granularities (e.g. words, n-grams, and sentences) on the domains of short stories, scientific abstracts, and song lyrics: 
\url{https://chrisdonahue.com/ilm}.
All code, data, and trained models are available at 
\url{https://github.com/chrisdonahue/ilm} and also on the CodaLab platform at  \url{https://worksheets.codalab.org/worksheets/0x9987b5d9cce74cf4b2a5f84b54ee447b}.

\section{Problem Statement}\label{sec:problem}
The task of infilling is to take incomplete text $\xtilde$, containing one or more missing spans, and return completed text $\x$. 
Let \blank{} be a placeholder for a contiguous sequence (span) of one or more missing tokens.
Then, incomplete text $\xtilde$ is a sequence of tokens some of which are \blank{}.
In order to map $\xtilde$ to $\x$, an infilling strategy must specify both \emph{how many} and \emph{which} tokens to generate for each \blank. 
Note that there may be many reasonable $\x$ for a given $\xtilde$. 
Hence, we are interested in learning a distribution $p(\x \mid \xtilde)$. 

\section{Infilling by Language Modeling}\label{sec:approach}
In this section, we describe our ILM framework. 
We first outline a simple reparametrization of the infilling task. 
Then, we define a procedure for automatically generating suitable training examples which can be fed to an off-the-shelf LM.

\subsection{Formulation}\label{sec:formulation}

\citet{fedus2018maskgan} explore an infilling framework where LMs are trained on concatenations of $\xtilde$ and $\x$, i.e., they use LMs to directly predict $\x$ given $\xtilde$. 
While their approach is effective at infilling individual words, 
it is somewhat redundant as the model must ``predict'' the unmasked text in $\xtilde$. 
Additionally, a model is not guaranteed to exactly reproduce the unmasked text.

Instead, 
we make the trivial observation that it suffices to predict only the missing spans $\y$ which will replace the \blank{} tokens in $\xtilde$. 
We can then construct $\x$ by simply replacing \blank{} tokens in $\xtilde$ with predicted spans $\y$ in a deterministic fashion. 
In order to handle multiple variable-length spans, we pose $\y$ as the concatenation of all missing spans separated by special \answer{} tokens (one \answer{} per \blank{}) (\Cref{fig:infilling_training_example}). 
We can thus cast infilling as learning $p(\y \mid \xtilde)$ without loss of generality.

\subsection{Training}\label{sec:training}

Given a corpus consisting of complete text examples,
our framework first manufactures \emph{infilling examples} 
and then trains an LM on these examples. 
To produce an infilling example for a given $\x$, 
we first sample an $\xtilde$ from a stochastic function $\Mask$ which randomly replaces some number of spans in $\x$ with \blank{} tokens. 
Then, we concatenate together the spans which were replaced---separated by \answer{} tokens---to form a training target $\y$.
Finally, we construct the complete infilling example by concatenating $\xtilde$, \sep{}, and $\y$ (see \Cref{fig:training_examples} for a complete example).

We train (or fine-tune) LMs on these infilling examples using standard LM training methodology, yielding models of the form $\p(\y \mid \xtilde)$. 
Specifically, we train GPT-2~\citep{radford2019language} off the shelf, but any LM can potentially be used. 

This framework has several advantages.
First, it incurs almost no computational overhead compared to language modeling. 
Specifically, if there are $k$ missing spans in $\xtilde$, 
the concatenation of $\xtilde$ and $\y$ contains only $2k+1$ more tokens than $\x$ (one \blank{} and one \answer{} per missing span plus one \sep{}).
As $k$ is usually small (averaging around $2$ per example in our experiments), 
sequence lengths remain similar to those encountered for the same $\x$ during language modeling. 
In contrast, using LMs to directly predict $\x$ from $\xtilde$ as in \citet{fedus2018maskgan} effectively doubles the sequence length of $\x$. 
This is particularly problematic when considering models like GPT-2 whose memory usage grows quadratically with sequence length. 
Second, our framework requires minimal change (three additional tokens) to an existing LM's vocabulary.
Finally, because the entirety of $\xtilde$ is in the ``past'' when predicting $\y$, the ILM framework combines the ability to attend to incorporate context on both sides of a blank with the simplicity of decoding from LMs. 

\section{Experimental Setup}\label{sec:experiments}
We design our experiments to determine if training an off-the-shelf LM architecture with our ILM framework can produce effective infilling models for a variety of datasets.
Specifically, we train on three datasets of different sizes and semantics (details in \Cref{sec:datasets}): short \stories~\citep{mostafazadeh2016corpus}, CS paper \abstracts{}, and song \lyrics{}.

\subsection{Mask Function}\label{sec:mask}

A benefit of the ILM framework is that it can be trained to infill spans corrupted by arbitrary mask functions.
Here, we explore a mask function which simultaneously trains models to infill different \emph{granularities} of text; specifically, words, n-grams, sentences, paragraphs, and documents. 
By using a unique special token per granularity (e.g. \blankword), 
this mask function offers users coarse but intuitive control over the length of the spans to be infilled.

We configure our mask function to mask each token in a given document with around $15$\% probability, echoing the configuration of \citet{devlin2019bert}. 
However, instead of masking individual tokens uniformly at random, 
we perform a pre-order traversal of the granularity hierarchy tree, randomly masking entire subtrees with $3$\% probability. 
For the datasets we consider, this results in a marginal token mask rate of about $15$\% (details in \Cref{sec:mask_deets}).

While we train to infill several different granularities, we primarily evaluate and discuss the ability of our models to infill sentences for brevity. 
Quantitative results of our models on other granularities can be found in~\Cref{sec:eval_granularities}, and granularity functionality can also be explored in our web demo.

\subsection{Task and Model Configurations}\label{sec:task_configs}

\begin{figure}[t]
    \centering
    \includegraphics[width=1\linewidth]{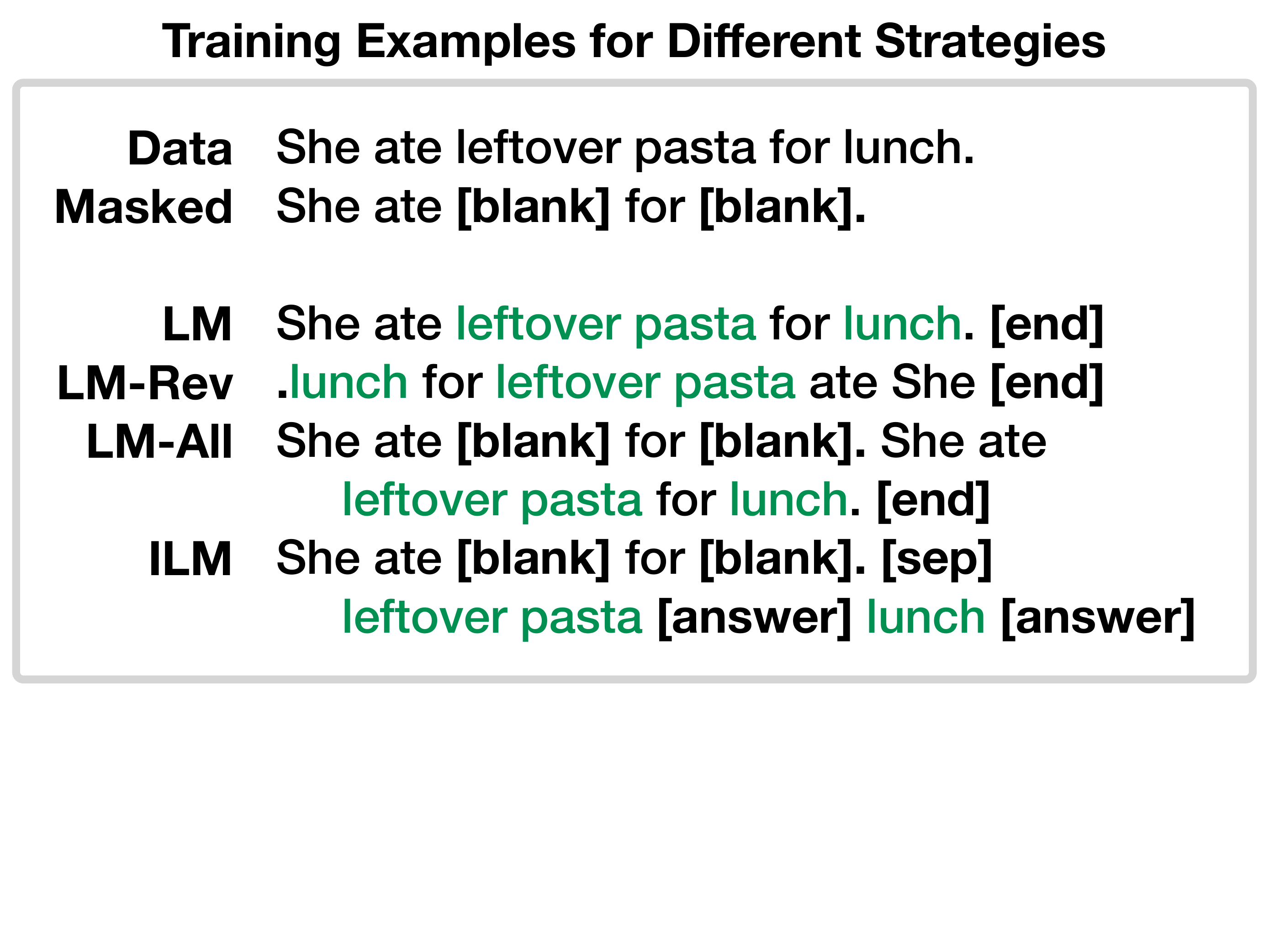}
    \vspace{-2cm}
    \caption{
    Training examples for three baseline infilling strategies and ILM on a given artificially-masked sentence. 
    For each strategy, we train the same architecture (GPT-2) on such examples. 
    At both training and test time, examples are fed from left to right; anything to the left of a green target is available to the model as context when predicting the target.
    Precisely, \lm{} only considers past context, and \lmrev{} only considers future. \lmall{} considers all available context but uses long sequence lengths. Our proposed \ilm{} considers all context while using fewer tokens. 
    }
    \label{fig:training_examples}
\end{figure}

For all experiments, we train the same architecture (GPT-2 ``small'') using the same hyperparameters (\Cref{sec:hyperparams}) while varying the infilling strategy and dataset. 
In addition to our proposed \ilm{} strategy for infilling, we consider three baseline strategies:
(1)~language modeling (\lm{}; ``infilling'' based only on past context),
(2)~reverse language modeling (\lmrev{}; ``infilling'' based only on future context),
and 
(3)~language modeling based on all available context (\lmall{}).
\lmall{} simply concatenates $\x$ and $\xtilde$ together as in~\citet{fedus2018maskgan}. 
\lmall{} represents arguably the simplest way one could conceive of infilling with LMs, 
but results in long sequence lengths.
Training examples for all strategies are depicted in \Cref{fig:training_examples}.

For each strategy, we also vary whether training is initialized from the pre-trained GPT-2 model or from scratch.
Despite discrepancies between the pre-training and our fine-tuning for most infilling strategies, 
\emph{all} of the infilling experiments initialized from the pre-trained checkpoint performed better than their from-scratch counterparts.
This indicates that ILM can effectively leverage large-scale language modeling pre-training to improve infilling performance. 
Henceforth, we will only discuss the models initialized from the pre-trained checkpoint, though we report quantitative performance for all models in \Cref{sec:eval_granularities}.

For the models trained on \stories{} and \abstracts{}, we trained models to convergence using early stopping based on the validation set perplexity (PPL) of each model computed only on the masked tokens. 
These models took about a day to reach their early stopping criteria on a single GPU. 
For the larger \lyrics{} dataset, we trained models for $2$ epochs (about two days on a single GPU). 

\section{Quantitative Evaluation}\label{sec:statistical}
\begin{table}[t]
    \centering
    \begin{tabular}[t]{lcccc}
        \toprule
            & \sto{}   & \abs{}   & \lyr{}   & Length \\
        \midrule
\lm{} & 18.3 & 27.9 & 27.7 & 1.00\\
\lmrev{} & 27.1 & 46.5 & 34.3 & 1.00 \\ 
\lmall{} & 15.6 & 22.3 & 21.4 & 1.81\\
\ilm{} & 15.6 & 22.4 & 22.6 & 1.01 \\
        \bottomrule
    \end{tabular}
    \caption{Quantitative evaluation results. We report test set perplexity (PPL) on the sentence infilling task for different model configurations on all three datasets, as well as average length of all test set examples in tokens relative to that of the original sequence (lower is better for all columns). 
    Our proposed \ilm{} framework achieves better PPL than both \lm{} and \lmrev{}, implying that it is able to take advantage of both past and future context. 
    \ilm{} achieves similar PPL to \lmall{} with shorter sequence lengths (hence less memory).}
    \label{tab:ppl_sentences}
\end{table}%

We evaluate the quantitative performance of our models on the sentence infilling task by measuring PPL on test data.\footnote{Overlap-based metrics such as BLEU score~\citep{papineni2002bleu} are not appropriate for evaluating infilling as there are many realistic infills that have no word-level overlap with the original, e.g., ``a sandwich'' instead of ``leftover pasta.''} 
In this setting, a sentence is selected at random and masked out, and we measure the likelihood assigned by a model to the masked sentence in the context of the rest of the document. 
Regardless of differences in the ordering and number of tokens that each strategy uses to represent a test example, 
PPL is always computed only for the span of tokens comprising the original sentence (e.g. green tokens in \Cref{fig:training_examples}). 

\Cref{tab:ppl_sentences} shows that across all datasets, \ilm{} outperforms models which see only past or future context (\lm{} and \lmrev{} respectively), 
implying that our proposed framework is able to take advantage of bidirectional context despite using unidirectional models. 
Additionally, 
while one might expect \lmall{} to outperform \ilm{} because its training examples more closely ``resemble'' those of standard LMs, 
\ilm{} achieves similar performance to \lmall{}.
This indicates that GPT-2 is able to effectively learn the ``syntax'' of \ilm{} examples and achieve reasonable infilling performance with shorter sequences (and hence with much less memory usage).

We also observe that models trained via \ilm{} perform similarly on the special case of language modeling compared to the models which were trained \emph{only} on language modeling 
(\Cref{sec:quant_lm}). 
This suggests that ILM does not just repurpose LMs to infill, but rather \emph{extends} their capabilities while maintaining their original functionality.

\section{Human Evaluation}\label{sec:human}
In addition to our quantitative evaluation, we seek to evaluate the qualitative performance of ILM. 
To this end, we sample a story from the \stories{} test set and randomly replace one of its five human-written sentences with a model output. 
Then, we task human annotators on Amazon Mechanical Turk with identifying which of the sentences in a story was machine-generated (details in \Cref{sec:human_eval_details}).

We compare our \ilm{} model to three baseline infilling strategies: 
an \lm{} (context beyond the replaced sentence was discarded), 
the best model (self-attention; SA) from~\citet{zhu2019text},
and the pre-trained BERT (base) model~\citep{devlin2019bert}.
All approaches except for BERT were first fine-tuned on the \stories{} dataset. 
To infill using BERT, we replace the tokens representing the original sentence with mask tokens, and then generate text by replacing mask tokens one at a time (conditioning on previously-generated tokens). 
While vocabulary differences make it is less useful to compare PPL for the SA and BERT baselines to our GPT-2-based strategies, 
we can still meaningfully compare them in this human evaluation setting.

\begin{figure}[t]
    \centering
    \includegraphics[width=1\linewidth]{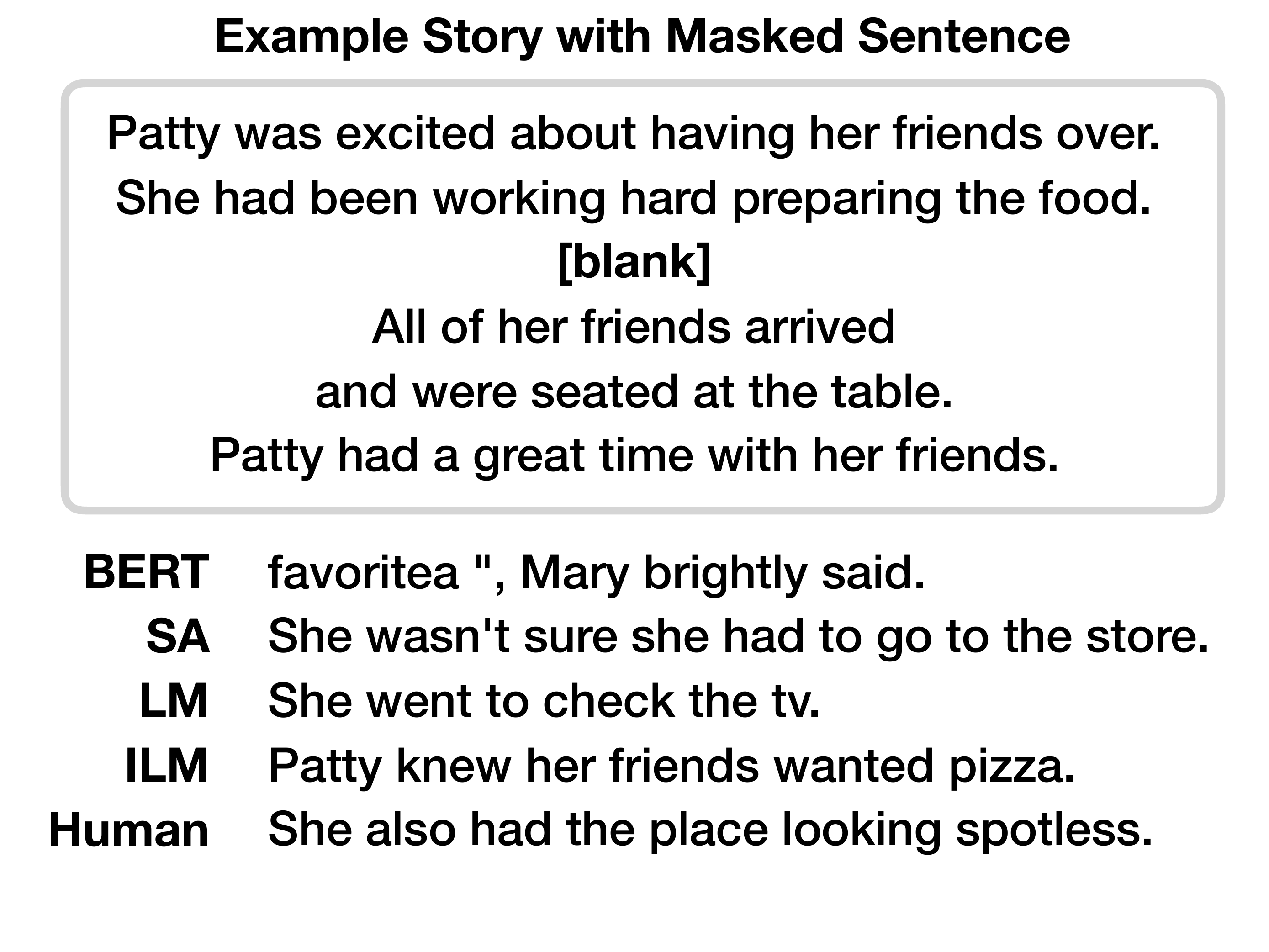}
    \vspace{-1cm}
    \caption{Example of a short story in our \stories{} dataset with its third sentence masked, and sentences infilled by different models. The sentences generated by BERT and SA models are off-topic, the sentence generated by LM model is irrelevant to the future context, while the ones generated by ILM and Human successfully account for both previous and future context.}
    \label{fig:human_eval_example}
\end{figure}

For each approach we compute a \emph{score}, 
which we define as the percentage of examples where the annotator did not correctly identify the machine-generated sentence. 
Therefore, a higher score implies a better (more natural, human-like) model.
We collect $100$ responses for each model and report the scores in \Cref{tab:humaneval}, with qualitative examples in \Cref{fig:human_eval_example} and \Cref{sec:more_examples}. 

\begin{table}[t]
    \centering
    \begin{tabular}[t]{ccccc}
        \toprule
        & BERT  & SA    & \lm{}  & \ilm{} \\
        \midrule
Score (\%)   & 20    & 29    & 41        & 45 \\
        \bottomrule
    \end{tabular}
    \caption{
Human evaluation results.
We use BERT~\citep{devlin2019bert}, the best model from~\citet{zhu2019text} (SA), and our \lm{} and \ilm{} models to replace random sentences in five-sentence stories from the \stories{} test set. 
Then, we task humans with identifying which sentence of the five was generated by a machine. 
We report the \emph{score} of each model: the percentage of infilled stories where the human failed to identify the machine-generated sentence. 
Our \ilm{} model achieves a higher score than all of the other models. 
Note that the max score is effectively 80\%, as a perfect model would cause annotators to randomly choose one of the five sentences.
}
    \label{tab:humaneval}
\end{table}

Of the four strategies, \ilm{} achieves the highest score, implying that sentences infilled by \ilm{} are harder for humans to recognize as fake than those produced by other strategies. 
Somewhat surprisingly, we observed that despite only observing past context the \lm{} model performed better than BERT and SA. 
BERT may have performed poorly due to the intrinsic difficulty of finding convincing infills with a precise length in tokens. 
SA may have performed poorly because, unlike \lm{} and \ilm{}, it was not initialized from a large-scaled pre-trained LM. 

\section{Related Work}\label{sec:related}
\paragraph{Methodology.}
A number of systems have the capability to infill but have practical drawbacks. 
Many systems are unable to automatically determine span length, and thus, can only infill fixed-length spans~\citep{fedus2018maskgan,devlin2019bert,yang2019xlnet,joshi2019spanbert,gu2019insertion,liu2019tigs}.
Methods such as BERT present additional challenges during inference~\citep{wang2019bert}.
\citet{rudinger2015script} frame narrative cloze as a generation task and employ language models, 
but they only consider one infill of a fixed length. 
\citet{zhu2019text,shen2020blank} infill multiple variable-length sequences, 
but these approaches require the masked context to be iteratively updated and reprocessed to fill in blanks one a time. 
In contrast, our approach appends infilled text to the context and does not require reprocessing the entire input sequence for each blank. 
\citet{ai2019haim} train an LM which can fill in the middle of a paragraph given the first and last sentences---our work generalizes to such capabilities.

\paragraph{Task.} The cloze task \citep{taylor1953cloze} evaluates language proficiency by asking systems to fill in randomly-deleted words by examining context. 
Cloze has been extended in the forms of discourse~\citep{deyes1984towards} and narrative cloze~\citep{chambers2008narrative}, 
which remove phrases and narrative events respectively.
Recently, 
cloze has been used not only for evaluation, but also to improve text generation quality~\citep{fedus2018maskgan}
and transfer learning~\citep{devlin2019bert} (under the name ``masked language modeling'').
Text infilling can be thought of as generalizing the cloze task from single words to spans of unknown length.
\citet{raffel2019exploring} explore infilling as a pre-training objective to improve downstream performance on inference tasks;
our work focuses on generation.

\paragraph{Story generation.}

Recent work seeks to generate stories given a title and storyline \citep{peng2019plan}, 
entities \citep{clark2018neural}, 
premise \citep{fan2018hierarchical}, 
or surrounding context and rare words \citep{ippolito2019unsupervised}. 
Our work differs in that we aim to build systems capable of making predictions based only on text context, 
rather than aspects specific to stories (e.g. storyline).

\section{Conclusion}\label{sec:conclusion}
We presented a simple strategy for the task of infilling which leverages language models. 
Our approach is capable of infilling sentences which humans have difficulty recognizing as machine-generated. 
Furthermore, 
we demonstrated that our infilling framework is effective when starting from large-scale pre-trained LMs, 
which may be useful in limited data settings. 
In future work, we plan to incorporate these features into co-creation systems which assist humans in the writing process. 
We hope that our work encourages more investigation of infilling, 
which may be a key missing element of current writing assistance tools.

\section*{Acknowledgments}

This work was funded by DARPA CwC under ARO prime contract no. W911NF-15-1-0462.
We thank all reviewers for their helpful comments.

\bibliography{all,acl2020}
\bibliographystyle{acl_natbib}

\clearpage

\appendix

\section{Datasets}\label{sec:datasets}

\begin{itemize}[leftmargin=*,label=-]
    \item \stories{} ($100$K examples, $5$M words) \\
    Short stories from the ROCStories dataset \citep{mostafazadeh2016corpus}. Each story contains a title and five sentences.
    \item \abstracts{} ($200$K examples, $30$M words) \\
    Abstracts from CS papers on arXiv
    \item \lyrics{} ($2$M examples, $60$M words) \\
    Song lyrics from \url{lyrics.com}
\end{itemize}

\noindent We experimented on multiple datasets to demonstrate that our framework was not custom tailored to a single domain. 
On the \stories{} and \abstracts{} datasets, we include metadata (story title, paper subject matter, etc.), as the first ``paragraph'' of the document. 
By providing these paragraphs (\Cref{sec:mask_deets}), our infilling model implicitly learns to summarize (e.g. infill a title given a story), and do conditional generation (e.g. infill a story given a title).
On the \lyrics{} dataset, infilling models may be especially helpful to humans; external aid in the form of rhyming dictionaries is already commonly employed in this domain.

To ensure that all experiments were trained on the same data,
we removed infilling examples which would have exceeded our training sequence length of $256$ tokens for the model with the longest sequence length (\lmall{}). 
This removed no examples from \stories{}, a small fraction of examples from \lyrics{}, and a substantial number of examples from \abstracts{}. 

\section{Masking function}\label{sec:mask_deets}

We design a mask function which takes the entire document and selectively masks several span granularities: words, $n$-grams, sentences, paragraphs, and entire documents. 
Accordingly, models trained via ILM on this masking function offer users the ability to specify the granularity of text to infill at a particular location. 
This allows users to have coarse but intuitive control over infilling length, so that multiple paragraphs are not generated when the user was expecting a single word.

Our masking function first constructs a tree of the training example (using the natural hierarchy of documents, paragraphs, sentences, and words). 
Then, using a pre-order tree traversal, each subtree is masked with $3\%$ probability (or ignored if any of its ancestors are already masked).
If the entire document (root node of the tree) is masked, then the infilling model's job is equivalent to that of a language model. 
If a word (leaf) is selected to be masked, $50\%$ of the time we mask that individual word, otherwise we mask an $n$-gram of random length between $1$ and $\min(8, \text{\# words left in the sentence})$ words (inclusive). 
Note that a word may comprise multiple tokens, as GPT-2 uses sub-word tokenization~\citep{sennrich2015neural}.
We chose the value of $3\%$ as, for the datasets we considered, it resulted in a marginal token mask rate of around $15\%$, echoing the configuration of~\citet{devlin2019bert}.

We add special tokens for each granularity to our model's vocabulary (e.g. \blankword), 
so that the user may specify which granularity they would like the infilling model to produce.
This functionality can be explored in our demo: \url{https://chrisdonahue.com/ilm}.

While we focus on this specific mask function in this paper, 
we structured the ILM codebase to allow users to train infilling models for completely different use cases. 
Users need only define a new mask function which takes complete documents and outputs lists of character-level spans representing the desired spans to be masked.

\section{Hyperparameters}\label{sec:hyperparams}

We use early stopping based on the PPL of the model on infilling the masked token for the validation set. 
We train all models using the default fine-tuning parameters specified in the \texttt{transformers} library~\citep{wolf2019transformers}, except that we use a batch size of $24$ and a sequence length of $256$. 

Note that the most straightforward way of training an LM on ILM examples (\Cref{sec:training}) is to maximize the likelihood of the entire concatenated example: $\xtilde$, $\text{\sep}$, and $\y$. 
This trains the model to predict tokens in $\xtilde$ even though such behavior is not necessary at inference time as $\xtilde$ will always be fully-specified. 
Nevertheless, we found that this additional supervision \emph{improved} performance when evaluating model PPL of $\y$. 
Conveniently, this is also the default behavior when adapting existing LM training code for use with ILM.

\section{Evaluation on language modeling and infilling other granularities}\label{sec:eval_granularities}

Our quantitative evaluation (\Cref{sec:statistical}) examined the sentence infilling performance of GPT-2 initialized from the large-scale pre-trained checkpoint after fine-tuning on different datasets and infilling strategies. 
Here, we report PPL for GPT-2 both initialized from scratch and from the pre-trained checkpoint for several other configurations: language modeling, a mixture of granularities, specific granularities, and language modeling.

\subsection{Language modeling}\label{sec:quant_lm}

\begin{table}[t]
    \centering
    \begin{tabular}[t]{lccc}
        \toprule
            & \sto{}   & \abs{}   & \lyr{} \\
        \midrule
\lmscratch{} & 33.4 & 52.1 & 25.1 \\
\lmrevscratch{} & 32.9 & 53.9 & 24.7 \\
\lmallscratch{} & 30.4 & 44.6 & 26.2 \\
\ilmscratch{} & 30.8 & 45.3 & 30.6 \\
\lm{} & 17.6 & 25.7 & 20.8 \\
\lmrev{} & 25.1 & 36.7 & 23.7 \\
\lmall{} & 17.8 & 25.2 & 21.5 \\
\ilm{} & 18.1 & 23.9 & 23.0 \\
        \bottomrule
    \end{tabular}
    \caption{Document infilling PPL (or language modeling) of ILM and baselines initialized either from scratch or from the pre-trained checkpoint across three datasets. Note that PPL of \ilm{} is similar to \lm{}, implying that our infilling strategy can reasonably maintain the ability to perform language modeling while extending the ability to infill.}
    \label{tab:granu_ppl_documents}
\end{table}

In~\Cref{tab:granu_ppl_documents}, we report PPL for ``document infilling,'' which is equivalent to language modeling (because $\xtilde$ is always \blankdocument{}). 
Because of how we structured our mask function (\Cref{sec:mask_deets}), 
$3$\% of infilling examples consist of the entire document masked out, 
which results in the ability of our ILM framework to perform standard infilling. 
We see that performance of \ilm{} is similar to that of \lm{} on this task, even though \ilm{} sees far fewer examples of language modeling compared to \lm{}.

\subsection{Mixture of granularities}

\begin{table}[t]
    \centering
    \begin{tabular}[t]{lccc}
        \toprule
            & \sto{}   & \abs{}   & \lyr{} \\
        \midrule
\lmscratch{} & 34.0 & 52.8 & 28.9 \\
\lmrevscratch{} & 34.9 & 59.3 & 30.4 \\
\lmallscratch{} & 27.0 & 46.2 & 24.3 \\
\ilmscratch{} & 25.5 & 46.0 & 27.5 \\
\lm{} & 17.5 & 25.5 & 23.9 \\
\lmrev{} & 26.5 & 39.0 & 29.2 \\
\lmall{} & 15.1 & 24.4 & 19.3 \\
\ilm{} & 14.9 & 23.5 & 20.2 \\
        \bottomrule
    \end{tabular}
    \caption{Mixture infilling PPL of all models (a mixture of all granularities).}
    \label{tab:granu_ppl_all}
\end{table}

In \Cref{tab:granu_ppl_all}, we report results for a mixture of granularities. 
Specifically, we run the same mask function we use for training (\Cref{sec:mask_deets}) on our test data and evaluate PPL on the masked spans.
This reflects general infilling ability across a wide variety of granularities (and hence lengths). 
Unlike our other quantitative evaluations, there may be multiple variable-length spans missing from each example in this evaluation.
Results are similar to that of sentence infilling. 
Namely, that \ilm{} outperforms \lm{} and \lmrev{} and is similar to \lmall{} despite using much less memory.

\subsection{Individual granularities}

\begin{table}[t]
    \centering
    \begin{tabular}[t]{lccc}
        \toprule
            & \sto{}   & \abs{}   & \lyr{} \\
        \midrule
\lmscratch{} & 35.6 & 51.5 & 25.1 \\
\lmrevscratch{} & 34.8 & 65.1 & 24.7 \\
\lmallscratch{} & 33.4 & 45.0 & 26.2 \\
\ilmscratch{} & 34.3 & 45.3 & 30.6 \\
\lm{} & 18.3 & 24.2 & 20.8 \\
\lmrev{} & 26.5 & 42.8 & 23.7 \\
\lmall{} & 20.4 & 23.4 & 21.5 \\
\ilm{} & 20.7 & 22.5 & 23.0 \\
        \bottomrule
    \end{tabular}
    \caption{
    Paragraph infilling PPL of all models.
    }
    \label{tab:granu_ppl_paragraphs}
\end{table}

\begin{table}[t]
    \centering
    \begin{tabular}[t]{lccc}
        \toprule
            & \sto{}   & \abs{}   & \lyr{} \\
        \midrule
\lmscratch{} & 36.0 & 65.4 & 33.5 \\
\lmrevscratch{} & 35.1 & 92.2 & 35.8 \\
\lmallscratch{} & 27.1 & 53.8 & 27.1 \\
\ilmscratch{} & 26.7 & 51.0 & 31.0 \\
\lm{} & 18.3 & 27.9 & 27.7 \\
\lmrev{} & 27.1 & 46.5 & 34.3 \\
\lmall{} & 15.6 & 22.3 & 21.4 \\
\ilm{} & 15.6 & 22.4 & 22.6 \\
        \bottomrule
    \end{tabular}
    \caption{Sentence infilling PPL of all models.}
    \label{tab:granu_ppl_sentences}
\end{table}

\begin{table}[t]
    \centering
    \begin{tabular}[t]{lccc}
        \toprule
            & \sto{}   & \abs{}   & \lyr{} \\
        \midrule
\lmscratch{} & 36.1 & 62.5 & 34.1 \\
\lmrevscratch{} & 36.4 & 89.1 & 36.3 \\
\lmallscratch{} & 26.4 & 60.1 & 24.3 \\
\ilmscratch{} & 23.1 & 49.5 & 26.3 \\
\lm{} & 19.2 & 25.5 & 28.2 \\
\lmrev{} & 26.6 & 45.0 & 34.8 \\
\lmall{} & 14.5 & 20.5 & 18.6 \\
\ilm{} & 13.8 & 21.5 & 18.8 \\
        \bottomrule
    \end{tabular}
    \caption{N-gram infilling PPL of all models.}
    \label{tab:granu_ppl_ngrams}
\end{table}

\begin{table}[t]
    \centering
    \begin{tabular}[t]{lccc}
        \toprule
            & \sto{}   & \abs{}   & \lyr{} \\
        \midrule
\lmscratch{} & 32.3 & 57.2 & 34.8 \\
\lmrevscratch{} & 31.6 & 100.0 & 36.7 \\
\lmallscratch{} & 12.6 & 51.8 & 12.5 \\
\ilmscratch{} & 9.2 & 37.9 & 12.2 \\
\lm{} & 17.1 & 23.0 & 28.7 \\
\lmrev{} & 24.1 & 45.0 & 35.1 \\
\lmall{} & 7.5 & 15.8 & 9.5 \\
\ilm{} & 5.4 & 14.2 & 8.5 \\
        \bottomrule
    \end{tabular}
    \caption{Word infilling PPL of all models.}
    \label{tab:granu_ppl_words}
\end{table}

In \Cref{tab:granu_ppl_paragraphs,tab:granu_ppl_sentences,tab:granu_ppl_ngrams,tab:granu_ppl_words} we report PPL values for infilling performance on paragraphs, sentences, n-grams, and words, respectively, across the three datasets. 

For each granularity, we create one infilling example per document from the test set with exactly one masked span (randomly chosen from all spans of that granularity for that document).
Then, we compute PPL only on the tokens which comprise the masked span, i.e., PPL is computed for all models on exactly the same set of tokens.
Across all granularities, we observe that \ilm{} outperforms \lm{} and \lmrev{} and either outperforms or is comparable with \lmall{} while using less memory.

\section{Details on human evaluation}\label{sec:human_eval_details}\label{sec:more_examples}

For human evaluation, we sampled 100 stories from the test set of the \stories{} dataset. 
From each story, we masked out one sentence at a time, thereby resulting in 500 stories with masked sentences. 
Then we used these stories as context and tasked each model with infilling the masked sentence. 

We compared 8 models in total. 
In addition to the four models reported in \Cref{sec:human} (BERT, SA, LM, and ILM), we included the models which are initialized from scratch (as opposed to initialized from the large-scale pre-trained checkpoint) for exhaustive comparison.
Furthermore, to filter out spam, we used a control model which always generates ``This sentence was generated by a computer.''
Lastly, we included the original sentence from the dataset as a reference model (Human) to sanity check the max score is around 80\%.

\begin{figure}[ht]
    \centering
    \includegraphics[width=1\linewidth]{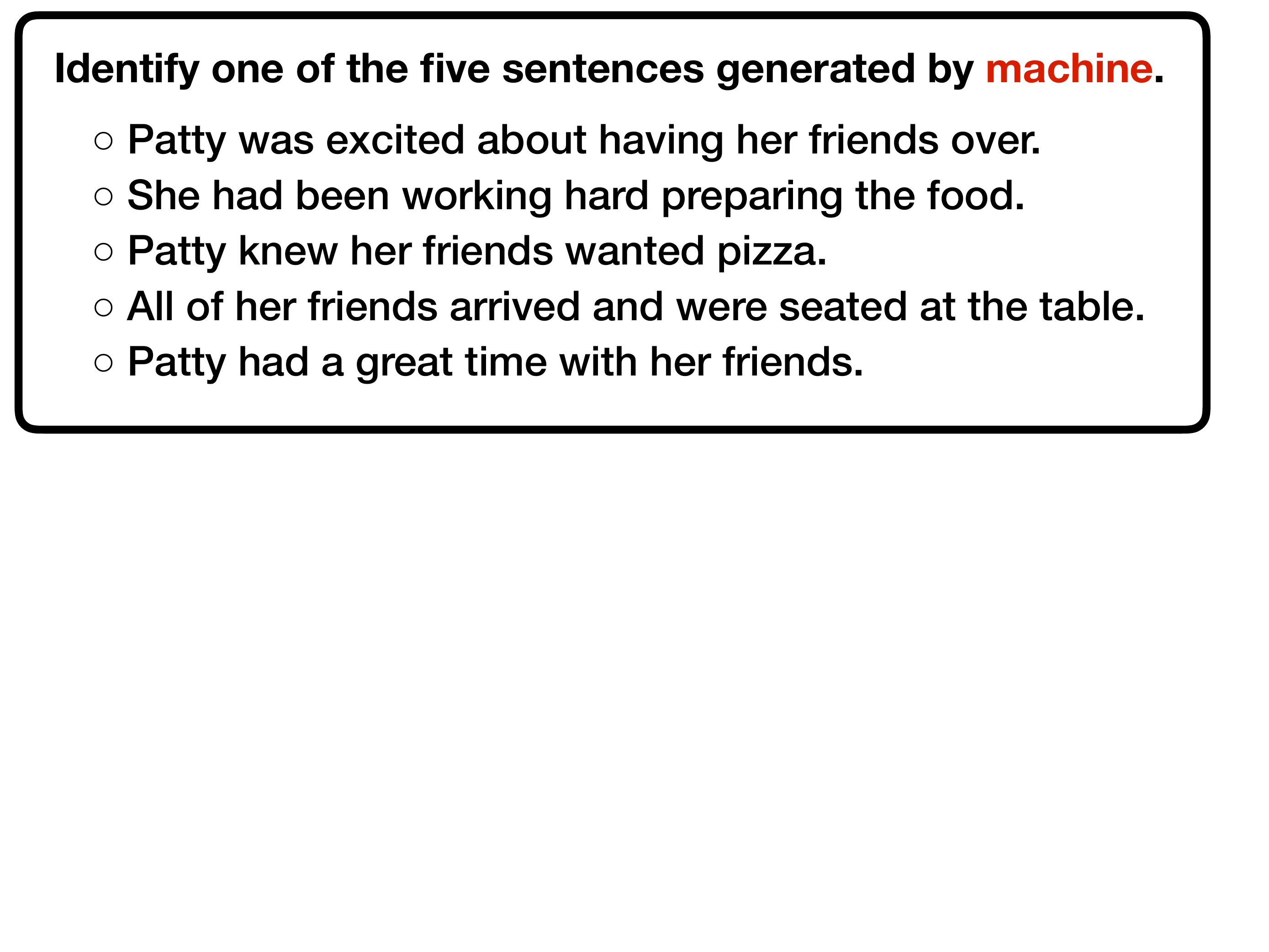}
    \vspace{-3.5cm}
    \caption{Example of a task and instruction for human evaluation on Amazon Mechanical Turk.}\label{fig:amt_example}
\end{figure}

\begin{figure}[H]
\includegraphics[width=1\linewidth]{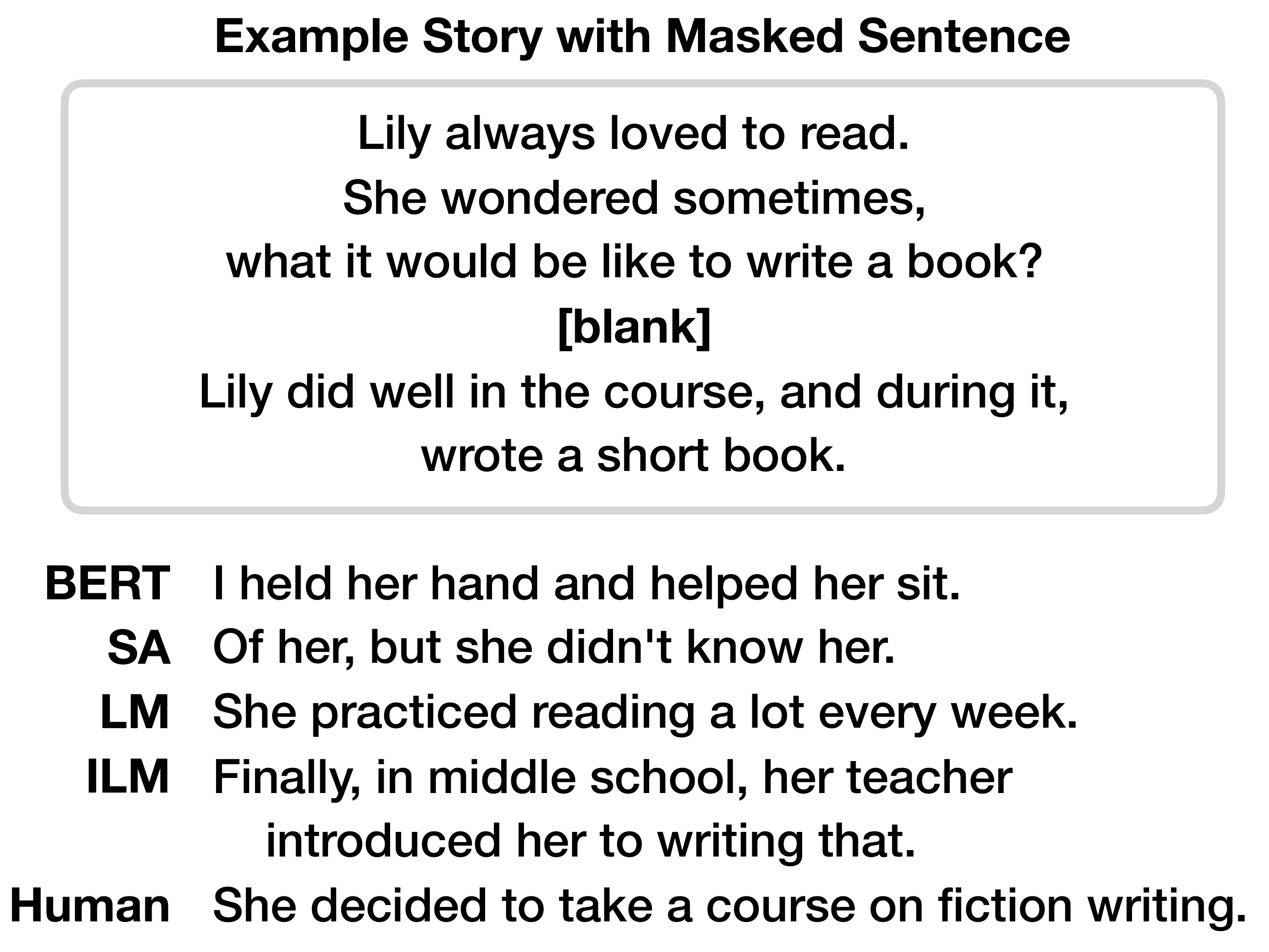}
\includegraphics[width=1\linewidth]{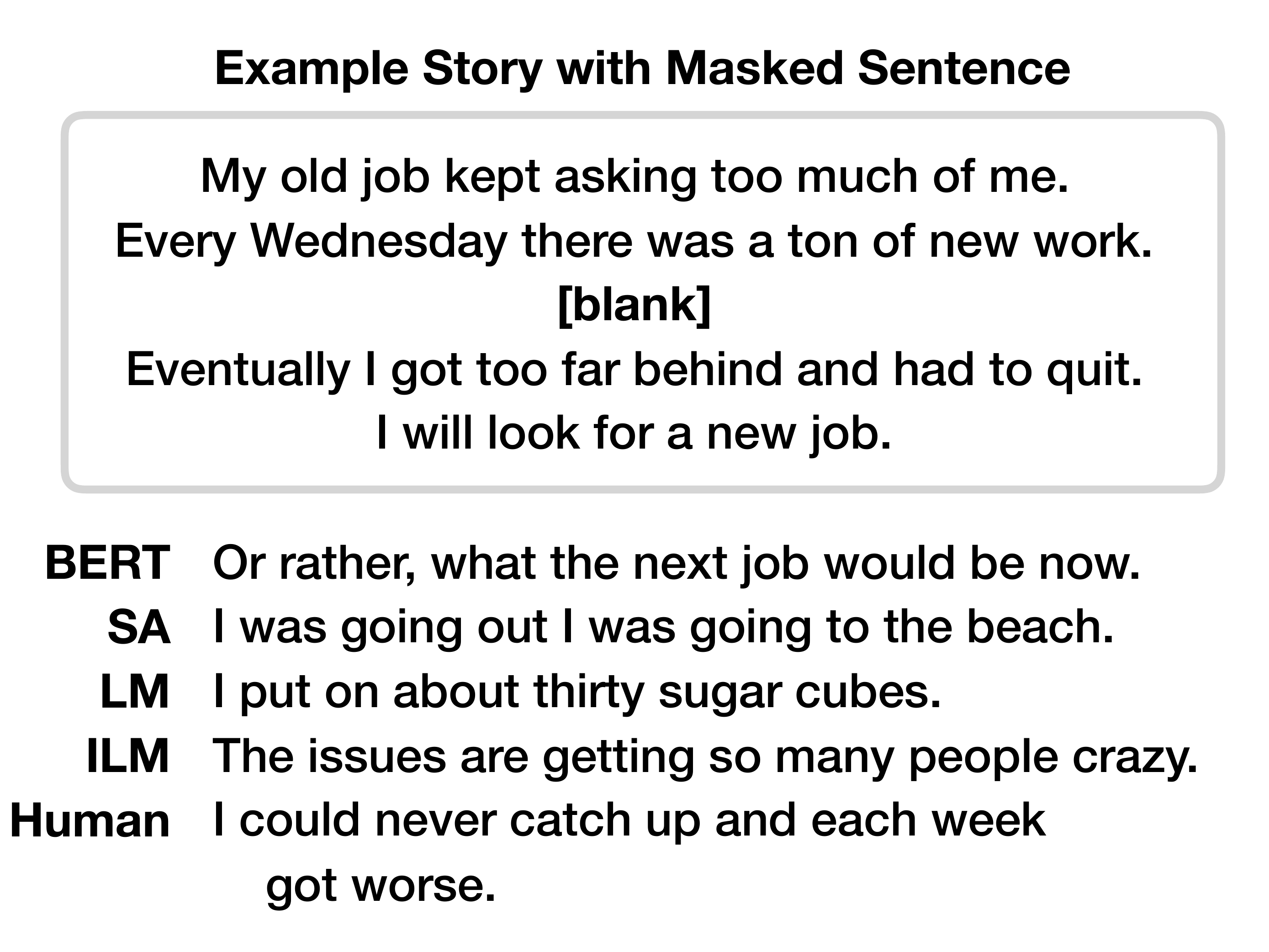}
\includegraphics[width=1\linewidth]{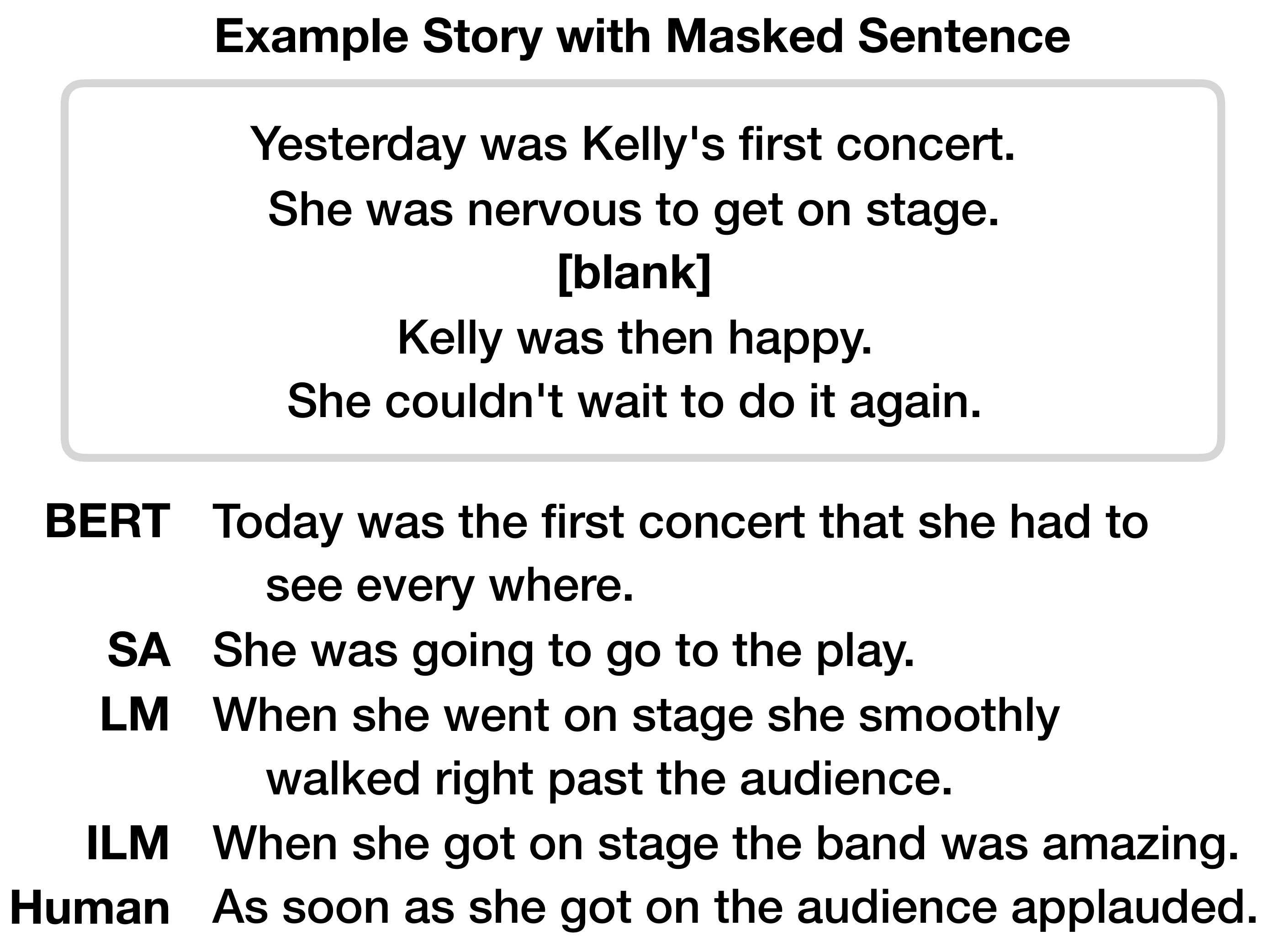}
\caption{Examples of sentence-level infills by different models.}\label{fig:more_examples}
\end{figure}

Each annotator was shown 8 stories, one from each model, and was asked to identify one of the five sentences generated by machine (see \Cref{fig:amt_example} for an example). 
Among the 100 collected responses, we filtered out 5 responses whose annotation for the control model was wrong. 
The quantitative and qualitative results can be found in \Cref{tab:human_eval_all} and \Cref{fig:more_examples}, respectively.
All model outputs and responses of human evaluation can be found at \url{https://github.com/chrisdonahue/ilm}.

\begin{table}[ht]
    \centering
    \begin{tabular}[h]{lc}
        \toprule
         & Score (\%) \\
        \midrule
        Control & 0 \\
        BERT & 20 \\
        SA & 29 \\
        LM (scratch) & 40 \\
        LM & 41 \\
        ILM (scratch) & 39 \\
        ILM & 45 \\
        Human & 78 \\
        \bottomrule
    \end{tabular}
    \caption{Human evaluation results.}\label{tab:human_eval_all}
\end{table}

\clearpage

\end{document}